\documentclass[journal]{IEEEtran}
\ifCLASSINFOpdf
\else
\fi
\usepackage{amsmath}
\usepackage{amsmath, amsfonts}
\usepackage{mathabx}   
\usepackage[makeroom]{cancel}    
\usepackage{tipa}      
\usepackage{textcomp}  
\usepackage{latexsym}  
\usepackage{mathrsfs}  

\usepackage{graphicx}
\usepackage{tikz}
\usepackage{pgfplots}
\pgfplotsset{compat=1.6}
\usepgfplotslibrary{groupplots}
\usepackage{pgf}
\usetikzlibrary{positioning,shapes,arrows,fit,calc,automata,perspective,3d}
\usepackage{neuralnetwork}
\usepackage{tikz-network}

\usepackage{hyperref}
\hypersetup{
	colorlinks=true,       
	linkcolor=blue,
	citecolor=red,
	filecolor=magenta,      
	urlcolor=cyan           
}
\usepackage{cleveref}

\usepackage{hhline}

\usepackage{multirow}
\usepackage{colortbl} 

\usepackage{algpseudocode}
\usepackage{algorithm}

\usepackage{subcaption} 

\newtheorem{rem}{Remark}[section]

\newcommand{\TENS}[1]{\mathop{\mathbf{#1}}\nolimits} 
\newcommand{\pe}[1]{^{(#1)}}	

\newcommand{\N}{\ensuremath{\mathbb{N}}}    
\newcommand{\R}{\ensuremath{\mathbb{R}}}	

\begin{document}
%
\title{Automated Detection of Sport Highlights from Audio and Video Sources}
%
%
%

\author{Francesco Della Santa,~
        Morgana Lalli
\thanks{F. Della Santa is with the Dipartimento di Scienze Matematiche, Politecnico di Torino, Turin, Italy}
\thanks{F. Della Santa is with the Gruppo Nazionale per il Calcolo Scientifico INdAM, Rome, Italy}
\thanks{M. Lalli is with the Innovation Lab, Deltatre S.p.A.}
}

%
%

\markboth{Preprint, 2025}%
{Shell \MakeLowercase{\textit{et al.}}: Automated Detection of Sport Highlights from Audio and Video Sources}

%



\maketitle

\begin{abstract}
This study presents a novel Deep Learning-based and lightweight approach for the automated detection of sports highlights (HLs) from audio and video sources. HL detection is a key task in sports video analysis, traditionally requiring significant human effort. Our solution leverages Deep Learning (DL) models trained on relatively small datasets of audio Mel-spectrograms and grayscale video frames, achieving promising accuracy rates of 89\% and 83\% for audio and video detection, respectively. The use of small datasets, combined with simple architectures, demonstrates the practicality of our method for fast and cost-effective deployment. Furthermore, an ensemble model combining both modalities shows improved robustness against false positives and false negatives. The proposed methodology offers a scalable solution for automated HL detection across various types of sports video content, reducing the need for manual intervention. Future work will focus on enhancing model architectures and extending this approach to broader scene-detection tasks in media analysis.
\end{abstract}

\begin{IEEEkeywords}
Deep Learning, Neural Networks, Sport, Video Analysis, Highlight Detection, Convolutional Neural Network, Audio-Visual Models
\end{IEEEkeywords}

%
\IEEEpeerreviewmaketitle

\section{Introduction}\label{sec:intro}

\IEEEPARstart{T}{he} growing volume of sports content across various platforms has increased the demand for automated tools to detect key moments, such as goals, penalties, or other notable events, and compile them into highlights (HLs). This task is essential for creating game summaries, enhancing viewer engagement, and improving content recommendation systems. However, manually annotating sports videos is time-consuming and labor-intensive, making efficient and scalable automated solutions crucial for the sports media industry.

Traditionally, sports HL detection has relied on handcrafted features and rule-based approaches. For example, Xie et al. \cite{xie2004} proposed a framework combining domain knowledge and Hidden Markov Models (HMMs) to identify important events in soccer broadcasts. Similarly, Rui et al. \cite{rui2000} utilized game-specific heuristics, such as referee gestures and scoreboard overlays, to automatically extract potential HLs from baseball videos. These methods, while innovative for their time, required extensive manual tuning and lacked generalizability across different sports and video formats \cite{traditional_methods}.

Recent advancements in Deep Learning (DL) have revolutionized sports video analysis by enabling systems to learn meaningful patterns directly from data. For instance, Tran et al. \cite{tran2015} demonstrated the effectiveness of 3D Convolutional Neural Networks (3D CNNs) in capturing spatiotemporal features in video clips, which significantly improved action recognition tasks. Similarly, in \cite{Rongved2021_3dCNN_SoccerEvents}, Rongved et al. developed a 3D CNN model for real-time sports event detection. However, these approaches are computationally expensive and often require large, labeled datasets to achieve optimal performance. Alternative methods, such as Channel-Separated Convolutional Networks (CSNs), attempt to reduce computational complexity by separating channel interactions from spatiotemporal interactions \cite{csn2020}.

In parallel, the integration of audio data has been explored to enhance sports action detection. Vanderplaetse and Dupont \cite{vanderplaetse2020} demonstrated that combining audio and video streams improved action spotting accuracy by capturing crowd reactions and commentator speech. While their work highlights the importance of multi-modal data fusion, their method processes raw audio streams, potentially missing finer nuances such as tonal variations in human voices.

In this context, we propose a dual-stream approach that leverages both video and in-stadium audio to improve sports highlight detection. Our method takes a simplified, lightweight approach to video analysis by using 2D Convolutional Neural Networks (CNNs) to focus on spatial feature extraction from video frames, reducing model complexity compared to 3D CNNs. The audio stream is transformed into Mel-spectrograms, a technique that captures human voice frequencies, allowing the model to detect commentators' and fans' reactions, which are strong indicators of key moments in sports events.

Unlike previous work, our approach aims to achieve high accuracy with small datasets while maintaining computational efficiency. The proposed ensemble model combines predictions from both the video and audio streams, resulting in improved robustness against false positives and false negatives, making it adaptable to various sports and real-world scenarios.

Our contributions can be summarized as follows:
\begin{itemize}
    \item We propose an efficient and lightweight video analysis method using 2D CNNs, achieving accurate highlight detection without the need for computationally expensive 3D models.
    \item We utilize Mel-spectrograms to process game audio, focusing on human voice frequencies to capture commentators' and fans' reactions, which are strong indicators of key moments.
    \item We introduce a multi-modal ensemble model that combines predictions from both audio and video streams, improving detection accuracy and robustness.
    \item We demonstrate that our approach is scalable across various sports, showcasing examples applied to football where it achieves high accuracy with relatively small datasets.
\end{itemize}

The paper is organized as follows: \Cref{sec:problem} formulates the highlight detection problem and describes the datasets used in this study. \Cref{sec:modelarchs} outlines the model architectures, while \Cref{sec:training} details the training procedures and performance evaluation. Section \ref{sec:applications} describes how to use the model and presents experimental results. In the end, Section \ref{sec:conclusion} concludes the paper with a discussion of the work and of future directions.

\section{Problem Formulation and Datasets Creation}\label{sec:problem}

For developing Deep Learning (DL) models able to detect a chosen type of scene, from a given audio or video source, it is important to define the mathematical formulation of the scene-detection problem. Indeed, the problem formulation characterizes both the criteria for the dataset creation (see the next subsections), the model architecture (see \Cref{sec:modelarchs}), and the model applications (see \Cref{sec:applications}). 

In the following, we assume that the audio and video sources are recordings of professional football matches, while the type of scene we want to detect is the match highlights (HLs); of course, the same formulation can be generalized to any other kind of audio/video source (other sports, movies, etc.) and any other type of scene.

\subsection{Problem formulation}

Let $A$ and $V$ denote the sets of all the audio and video recordings of football matches, respectively. For each audio record $a\in A$ and each video record $v\in V$, we denote by $H(a)$ and $H(v)$ the sets of time intervals (in seconds) that identify the match highlights contained into $a$ and $v$, respectively. More precisely:
\begin{equation*}\label{eq:HLset}
    H(s):= \left\lbrace
    [t_{\rm start}^{(1)}, t_{\rm end}^{(1)}], \ldots , [t_{\rm start}^{(N_s)}, t_{\rm end}^{(N_s)}] 
    \right\rbrace
    \subset \N\,,
\end{equation*}
where $s$ is the type of recording (i.e., $s=a$ or $s=v$), and where $t_{\rm start}^{(i)}, t_{\rm end}^{(i)}\in\N$ denote the starting second and the ending second of the $i$-th highlight, respectively; $N_s\in\N$ denotes the total number of highlights in the recording $s$. Moreover, we have $t_{\rm start}^{(i)} < t_{\rm end}^{(i)}$, for each $i=1,\ldots, N_s$, and $t_{\rm end}^{(i)} < t_{\rm start}^{(i+1)}$ for each $i=1, \ldots, N_s-1$; indeed,  we assume that highlights cannot have a length less than one second and we assume that they do not overlap.

Now, let us define a \emph{highlight detection function} for audio chunks of $k$ seconds; i.e., we define the function $\mathcal{A}_k: A|_k\subset A \rightarrow \{0, 1\}$, for a fixed $k\in \N$, such that
\begin{equation}\label{eq:detect_audio}
    \mathcal{A}_k(a) = 
    \begin{cases}
        0 \quad & \text{if } H(a)=\emptyset\\
        1 \quad & \text{otherwise}
    \end{cases}
    \qquad 
    ,
\end{equation}
where $A|_k:=\{a\in A \ | \ a\text{'s length is }k \text{ seconds}\}$. 

Analogously, we define a \emph{highlight detection function} for video chunks $\mathcal{V}_k: V|_k\subset V \rightarrow \{0, 1\}$, for a fixed $k\in \N$, such that
\begin{equation}\label{eq:detect_video}
    \mathcal{V}_k(v) = 
    \begin{cases}
        0 \quad & \text{if } H(v)=\emptyset\\
        1 \quad & \text{otherwise}
    \end{cases}
    \qquad 
    ,
\end{equation}
where $V|_k:=\{v\in V \ | \ v\text{'s length is }k \text{ seconds}\}$. 

It is easy to observe that the functions $\mathcal{A}_k$ and $\mathcal{V}_k$ can be correctly evaluated for a chunk only if the chunk's seconds are labeled to know if a highlight is happening or not. However, in real applications, we want to detect highlights inside unlabeled audio/video sources; for this reason, we train DL models with the task of approximating \eqref{eq:detect_audio} and \eqref{eq:detect_video}, using sets of labeled source chunks of length $k$.

In the next two subsections, we will describe how we build the audio and video datasets necessary for training the DL models, given a fixed size $k\in\N$ for the source chunks.

\noindent\begin{rem}[On the choice of $k$ for building the datasets]\label{rem:k_effects}
    We observe that the smaller $k$, the more precise $\mathcal{A}_k$ and $\mathcal{V}_k$ are in detecting the highlights; i.e., if $k$ is small, more probably we have only one highlight in the chunk and, if $k$ is very small, the chunk can even coincide with a highlight-chunk. Then, the shorter the labeled chunks, the more precise the datasets. Nonetheless, the smaller $k$, the harder the training of a DL model in learning the detection task (because of less input information). Therefore, a careful choice of $k$ is important for finding a good trade-off between the detection precision of the target function and the difficulty in training the DL model for learning it.
\end{rem}

\subsection{Audio Source Dataset}\label{sec:audiodata}

\begin{figure}[htb]
    \centering
    \includegraphics[width=0.4\textwidth]{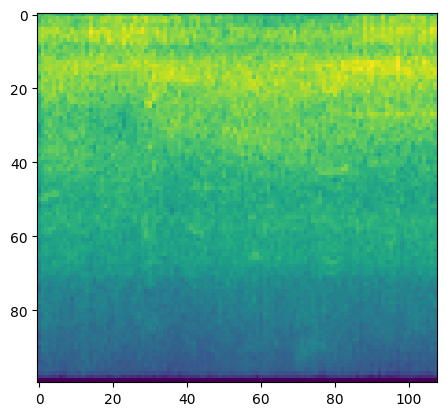}
    \caption{Example of Mel-spectrogram of an audio chunk}
    \label{fig:melspect}
\end{figure}

In DL, a good approach for working with audio sources is to transform them into Mel-spectrograms (e.g., see \cite{NIPS2013_b3ba8f1b,AKCAY202056,10154046,Nguyen2023,Latif2023}). These spectrograms are computed with respect to the Mel scale\cite{Stevens1940}, a scale useful to approximate the non-linear frequency response of the human ear; therefore, they are highly effective for analyzing human voices.
In particular, the Mel-spectrograms are processed as images (see \Cref{fig:melspect}), exploiting two-dimensional Convolutional Layers (CLs) \cite{Goodfellow2016_book}. For this reason, concerning the highlight detection task for audio sources, we build a dataset
\begin{equation*}
    \mathcal{D}_{\mathcal{A}_k}:=\left\lbrace (X_1, y_1),\ldots ,(X_D, y_D) \right\rbrace
\end{equation*}
such that $X_i\in\R^{p\times q}$ is the Mel-spectrogram of $a_i\in\ A|_k$ and $y_i=\mathcal{A}_k(a_i)$, for each $i=1,\ldots , D$. 

The $D\in\N$ samples of audio chunks $a_i\in A|_k$ are taken by splitting a set of available audio recordings $a\in A$ with known $H(a)$ and are selected in order to obtain a balanced dataset; i.e., $50\%$ of data have label $y_i=1$ and the other $50\%$ have label $y_i=0$, $i\in\{1,\ldots ,D\}$.

\subsection{Video Source Dataset}\label{sec:videodata}
In DL, the typical approach for working with colored video inputs is to stack the video frames (RGB images, i.e., three-dimensional tensors) into four-dimensional tensors and process them through three-dimensional CLs \cite{Ji2013_Conv3D,Tran2015_Conv3D}. 
Such an approach is efficient but also very expensive, both in terms of computational costs and computational resources for storing the DL model and the data.

\begin{figure}[htb]
    \centering
    \includegraphics[width=0.45\textwidth]{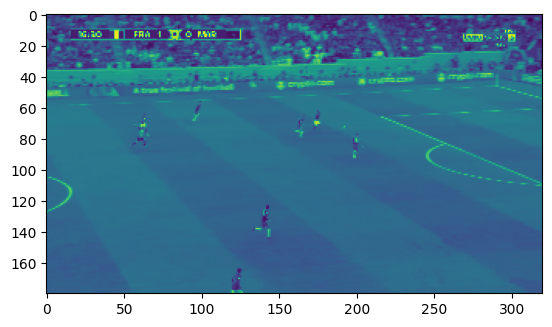}
    \caption{Example of black-and-white frame of a video frame}
    \label{fig:bw_frame}
\end{figure}

For performing our task (highlight detection) from video sources, we observe that colors are not so meaningful. Then, we can work with video chunks made of black-and-white frames (i.e., matrices of grey-scale pixels) stacked into three-dimensional tensors, representing the (black-and-white) video as an image with one channel per frame (see \Cref{fig:bw_frame}); therefore, we can process these video chunks through two-dimensional CLs. This approach permits the reduction of the memory dedicated to the data and to work with simpler DL models,  made of two-dimensional CLs instead of three-dimensional ones. Given this approach, we build a dataset
\begin{equation*}
    \mathcal{D}_{\mathcal{V}_k}:=\left\lbrace (\TENS{X}_1, z_1),\ldots ,(\TENS{X}_E, z_E) \right\rbrace
\end{equation*}
such that $\TENS{X}_i\in\R^{u\times w\times \varphi k}$ is the tensor representation of the black-and-white video chunk $v_i\in\ V|_k$, made of $\varphi k$ frames ($\varphi\in\N$ is the frame-rate) of size $u$-by-$w$ pixels. The labels are such that $z_i=\mathcal{V}_k(v_i)$, for each $i=1,\ldots , E$.

The $E\in\N$ samples of video chunks $v_i\in V|_k$ are taken by splitting a set of available video recordings $v\in V$ with known $H(v)$ and are selected in order to obtain a balanced dataset; i.e., $50\%$ of data have label $z_i=1$ and the other $50\%$ have label $z_i=0$, $i\in\{1,\ldots ,E\}$.

\section{Model Architectures}\label{sec:modelarchs}

In this section, we describe the general idea behind the architectures used for the two DL models trained for learning the highlights detection task from audio chunks and video chunks, respectively, of length $k$ seconds.

\noindent\begin{rem}[NDA and limited description of NN architectures]\label{rem:NDAarchs}
    The models described in this paper have been developed inside the Innovation Lab of the company Deltatre S.p.A., and the details of the NN architectures cannot be shared, because they are protected by a non-disclosure agreement (NDA). Nonetheless, we can report a general description of the models, to understand the fundamental characteristics of the NNs.
\end{rem}

\subsection{Model for Audio Sources}
The architecture for the DL model that is dedicated to learning the highlight detection task from audio chunks is conceptually simple. Indeed, since the audio chunks have been encoded into Mel-spectrograms (see \Cref{sec:audiodata}), we can build the model only using two-dimensional CLs, Pooling Layers (PLs), and Fully-Connected (FC) layers. 
More clearly, we can divide the architecture into two main blocks: a first block consisting of a series of CLs and PLs and a second block that is a series of FC layers. At the end of the second block, we have a last FC layer made of just one unit with \emph{sigmoid} activation function, because our target is a binary classification task (see \eqref{eq:detect_audio}).

\subsection{Model for Video Sources}
The architecture for the DL model dedicated to learning the highlight detection task from video chunks is strongly based on transfer learning techniques \cite{Tan2018_transfer_survey}. Indeed, for this model, we can exploit the large number of high-performance trained models for image classification that are available in the literature or accessible in private repositories (e.g., in the case of a company).

We point the reader's attention to the fact that we base our model on an architecture used for image classification tasks, not on an architecture for video classification tasks. The reasons for this choice are two:
\begin{enumerate}
    \item Our video chunks are encoded as $u$-by-$w$ images of $\varphi k$ channels, where each channel denotes a black-and-white video frame (see \Cref{sec:videodata}). On the contrary, video classification models typically work with 4-dimensional tensors where the fourth axis denotes the video frames (that can be either black-and-white or RGB images);

    \item Image classification models, usually based on 2-dimensional CLs, PLs, and FC layers only, typically are lighter than video classification models, which can involve also 3-dimensional CLs.
\end{enumerate}

Let us denote with $\mathcal{M}$ the trained model for image classification that we use for transfer learning; we assume that the model is built for working with RGB images of the same height and width as our video frames (i.e., $u$-by-$w$-by-$3$ input tensors) and that it performs a multi-class classification task. The operations that we apply to the model $\mathcal{M}$ for obtaining a model suitable for our video chunk classification task are listed in the following:
\begin{itemize}
    \item \textbf{Adaptation to input video chunks:} since $\mathcal{M}$ has been built for working with RGB images, we must modify its first 2-dimensional CL for working with our video chunks, without wasting the information contained in its kernel. Therefore, for each filter $K\pe{1},\ldots, K\pe{L} \in\R^{m\times n\times 3}$ of this CL, we compute $L$ new filters $\widetilde{K}\pe{1},\ldots ,\widetilde{K}\pe{L}\in\R^{m\times n\times \varphi k}$ such that the weights in each channel of the new filter $\widetilde{K}\pe{\ell}$ are the same, and equal to the mean of the weights of $K\pe{\ell}$ along its depth; i.e., for each $\ell=1,\ldots ,L$, and each $\kappa=1,\ldots ,\varphi k$, we have that
    \begin{equation}\label{eq:aggregated_filters}
        \left(\widetilde{K}\pe{\ell}\right)_{i j \kappa} := \frac{1}{3}\sum_{h=1}^3 \left(K\pe{\ell}\right)_{i j h}\,.
    \end{equation}
    The meaning of \eqref{eq:aggregated_filters} is that, for each fixed $\kappa$ and $\ell$, we have a one-channel filter that aggregates the information of the original filter for working with black-and-white images. Therefore, building the filter $\widetilde{K}\pe{\ell}$ by stacking $\varphi k$ times these one-channel filters, we obtain a filter that applies the knowledge learned by $\mathcal{M}$ to each frame of the input video chunk.

    \item \textbf{Adaptation to binary classification task:} since $\mathcal{M}$ has been built for a multi-class classification task, we substitute its output FC layer (with softmax activation function) with a FC layer made of one unit and with sigmoid activation function.

    \item \textbf{Setting trainable and fixed weights:} after the modification of the first CL layer and the substitution of the output layer, we must specify which layer weights are trainable and which are not. In a model like $\mathcal{M}$, at the end of the training, the block of CLs and PLs typically results in an encoder for images, that returns an encoded version of the information inside the pictures; then, the block of FC layers process the encoded images for classifying them. For these reasons, we will preserve the image encoding abilities of the block of CLs and PLs by freezing its layer weights (i.e., they are not trainable); on the other hand, since we perform a new classification task, we forget the values of the FC layers' weights, training them on our data from scratch.
\end{itemize}

\section{Training and Performance Evaluation of the Models}\label{sec:training}

In this section, we briefly describe the training procedure of both models introduced in \Cref{sec:modelarchs}. Then, we report their performance with respect to a test set.

\noindent\begin{rem}[NDA and limited description of NN training options]\label{rem:NDAtrain}
    As already discussed in \Cref{rem:NDAarchs}, the models described in this paper have been developed inside the Innovation Lab of the company Deltatre S.p.A. and they are protected by an NDA. Therefore, we cannot describe the details of the training procedures executed for obtaining the final, trained models. In particular, we will anonymize any quantity to hide their original values without losing the possibility of describing the procedure and the results.
\end{rem}

\subsection{Training of the Models}

Given a dataset $\mathcal{D}_{\mathcal{A}_k}$ of labeled audio chunks of $k$ seconds (encoded as Mel-spectrograms, see \Cref{sec:audiodata}) and a dataset $\mathcal{D}_{\mathcal{V}_k}$ of labeled video chunks of $k$ seconds (encoded as stacked black-and-white images, see \Cref{sec:videodata}), $k=5$ fixed, we randomly split each of them into training set, validation set, and test set. We recall that the chunks have been selected from highlight-labeled football match videos and are such that the datasets are balanced with respect to the labels; therefore, we can assume that the training, validation, and test sets are balanced, too. In order to give an idea of the size of the datasets used, we have that the cardinality of $\mathcal{D}_{\mathcal{A}_k}$ is made of approximately $700$ audio chunk samples, while $\mathcal{D}_{\mathcal{V}_k}$ is made of of approximately $1000$ video chunk samples.

Concerning the training of the models, we do not adopt strategies for searching optimal hyper-parameters and/or training options. We focus the training of each model on a single configuration of hyper-parameters and training options; these configurations have been selected after a brief preliminary investigation. We postpone to future work a deeper study, intending to further improve the performances reported in \Cref{sec:testset_perf}.

In the following list, we report a general description of the training configurations of the models:
\begin{itemize}
    \item \textbf{Audio model.} Dataset split: ${\sim}50\%$ training, ${\sim}10\%$ validation, ${\sim}40\%$ test; Minibatch size: ${\sim}4.5\%$ of the training set's cardinality; Regularization: early-stopping method.
    \item \textbf{Video model.} Dataset split: ${\sim}70\%$ training, ${\sim}10\%$ validation, ${\sim}20\%$ test; Minibatch size: ${\sim}1.2\%$ of the training set's cardinality; Regularization: early-stopping method.
\end{itemize}

\subsection{Performance of the Models}\label{sec:testset_perf}

The performances observed on the test sets are promising for both the audio model and the video model, with an accuracy of approximately $89\%$ and $83\%$, respectively. We also observe that the models have not only a good accuracy (that in these cases coincides with the recall index) but also good precision in detecting the chunks that intersect a highlight action (i.e., a low rate of false positives/negatives among predictions). Moreover, these performances are not unbalanced with respect to the positive or negative labels. See Tables \ref{tab:perf_summary}-\ref{tab:video_confmat} for more details about the models' performances.

\begin{table}[htb!]
    \centering
    \begin{tabular}{|c||c|c|c|}
        \hline
        Model & Accuracy/Recall & Precision & $F_1$-score\\
        \hline
        \hline
        Audio & 0.8924 & 0.8933 & 0.8923\\
        \hline
        Video & 0.8300 & 0.8308 & 0.8301\\
        \hline
    \end{tabular}
    \caption{Average performance indices for the binary classification task of audio/video chunks of length $k$ seconds on the corresponding test sets.}
    \label{tab:perf_summary}
\end{table}

\begin{table}[htb!]
    \centering
    \begin{tabular}{|c||c|c|}
        \hline
        LABELS & Pred. Pos. & Pred. Neg.\\
        \hline
        \hline
        True Pos. & ${\sim}43\%$ & ${\sim}7\%$ \\
        \hline
        True Neg. & ${\sim}4\%$ & ${\sim}46\%$ \\
        \hline
    \end{tabular}
    \caption{Audio model. Confusion matrix w.r.t. the test set. The values in the table have been normalized w.r.t. the cardinality of the test set (and represented as percentages). Perfect prediction performance would have value $50\%$ on the diagonal elements of the matrix.}
    \label{tab:audio_confmat}
\end{table}

\begin{table}[htb!]
    \centering
    \begin{tabular}{|c||c|c|}
        \hline
        LABELS & Pred. Pos. & Pred. Neg.\\
        \hline
        \hline
        True Pos. & $41\%$ & $9\%$ \\
        \hline
        True Neg. & $7\%$ & $43\%$ \\
        \hline
    \end{tabular}
    \caption{Video model. Confusion matrix w.r.t. the test set. The values in the table have been normalized w.r.t. the cardinality of the test set (and represented as percentages). Perfect prediction performance would have value $50\%$ on the diagonal elements of the matrix.}
    \label{tab:video_confmat}
\end{table}

\section{Application of the Models for Automated Highlights Detection}\label{sec:applications}

In this section, we illustrate the pipeline to automatically detect highlights in a video file using the models described in the previous sections. We recall that the advantage of using such kind of procedure, based on our DL models, is the notable reduction of human efforts required for performing a selection of highlight moments in a football match (see \Cref{sec:intro}).  

In a nutshell, given a video chunk of length $h$ seconds, $h\gg k$, we apply the models on a sliding window of $k$ seconds, estimating the possibility of having an intersection between the window and a highlight (the audio model is applied only to the audio file extracted from the video). Then, for each second of the video, we compute the average possibility of being part of a highlight, given all the estimates that the second received for each window that includes it. The detailed pipeline is illustrated in Subsection \ref{sec:pipeline} (items \ref{item:pipe_prepdata}-\ref{item:sourcespec_pred}); nonetheless, here we point the attention of the reader to the following details:
\begin{enumerate}
    \item \textbf{Continuous values for predictions:} we are not classifying the windows (and the seconds, as a consequence) with a label $1$ or $0$, but we use the rough output of the last layer of the model, characterized by a sigmoid activation function (i.e., values in $[0, 1]\subset\R$). Then, the average predicted score $\widehat{H}_s(\tau)\in[0, 1]$ for the second $\tau$ of the video file represents the possibility that $\tau$ belongs to a highlight time interval with respect to source $s$, $s\in\{a, v\}$.
    \item \textbf{Explanation of the predicted score's meaning:} as written in the item above, the average predicted score for the second $\tau$ represents the possibility for that second of belonging to a highlight time interval. Therefore, we can observe some ``\emph{unexpected-but-correct}'' scores in certain cases. For example, we can observe cases where a second $\tau$ with a large score corresponds to a moment of silence in the audio source, but only because after a few seconds (in the range of the $k$ seconds of the model's time window), a team suddenly scores a goal and the audio source ``explodes''. A similar example can be observed for the video sources; e.g., when we have large scores for a second $\tau$ corresponding to an apparently harmless action of a team that unexpectedly scores a goal a few seconds later.

    We want to emphasize that this kind of prediction is an advantage and not a disadvantage for the applications of the model. Indeed, a typical highlight does not focus on the single instant that characterized the scene (e.g., the instant of the goal), but it is a chunk that comprehends the start of the action that originated the key event and that continues with the reactions of players, commentators, and supporters to it.
\end{enumerate}

With a preliminary application of the models to the video files, using the pipeline described at the beginning of this section, we observed that each model may suffer from false positive/negative detections in some specific contexts. For example: 
\begin{itemize}
    \item \textbf{Audio false positives/negatives.} The audio model often recognizes as probable highlights those audio chunks where the spectators at the stadium and/or the commentators are particularly noisy. Therefore, for example, when chants last for a long period after a goal, the model can consider these chunks as chunks containing part of a highlight (false positive). On the other hand, the model can fail to detect chunks containing a highlight moment that is characterized by a not-so-loud noise; for example, when most of the spectators and the commentators are supporters of the team that concedes a goal.

    \item \textbf{Video false positives/negatives.} The video model often recognizes as probable intersections with highlights those video chunks where the camera is near the goal of a team, while rarely recognizes intersections with highlights for chunks where the camera is focused on spectators (or similar behaviors). Therefore, the video model may consider as a probable highlight intersection a chunk where the goalkeeper is alone and is kicking the ball (false positive example); on the other hand, the model may not consider as a highlight intersection a chunk where supporters exult after a goal of their team (false negative example).
\end{itemize}

From these observations, we noted that false positive/negative audio-based detections are typically compensated by true positive/negative video-based detections, and vice-versa. Therefore, we decided to build an ensemble model, based both on audio and video sources and both models for notably improving the prediction performances. In particular, the final estimate of highlight possibility for a second $\tau$ is the average between the audio average score and the video average score; in this way, the new ensemble model can reduce both false positive and false negative detections, with respect to the models that compose it (see \ref{item:ensembled_pred} in Subsection \ref{sec:pipeline}, below).

\subsection{Detailed Pipeline}\label{sec:pipeline}

Here, we illustrate in detail the pipeline used to automatically detect highlights from a video file, using the audio model, the video model, or the ensemble model (i.e., audio and video models together).
Let us denote by $\mathcal{M}_a$, $\mathcal{M}_v$, and $\mathcal{M}_e$ the audio, video, and ensemble models, respectively. Then, the pipeline consists of the following steps:
\begin{enumerate}
    \item\label{item:pipe_prepdata} Separate the audio source and the ``only-video'' source in the given video file; then, transform the video source into a black-and-white video (if the original file was in RGB colors) and adapt the frame-rate to $\varphi$ (frame-rate of $\mathcal{M}_v$).
    
    \item Split the sources in time windows of $k$ seconds and compute the Mel-spectrograms of each audio window. The overlap between the windows is arbitrary; obviously, the larger the overlap, the more precise the detection, and the more memory is occupied by data.
    
    \item For each source $s=a, v$ and for each $k$-seconds window $\omega$, estimate the highlight possibility rate with the model $\mathcal{M}_s$ (i.e., the raw output in $[0, 1]\subset \R$). We denote the highlight possibility rate of $\omega$, with respect to the source $s$, as $\mathcal{M}_s(s\pe{\omega})$; where $s\pe{\omega}$ denotes the $k$-seconds interval of source $s$ corresponding to $\omega$ and properly encoded for being a model's input.

    \item\label{item:sourcespec_pred} For each second $\tau$ in the time length of the video file, evaluate its average estimated highlight possibility rate assigned from the models $\mathcal{M}_a$ and $\mathcal{M}_v$; i.e., for each $s=a,v$, evaluate the quantity
    \begin{equation}
        \widehat{H}_s(\tau) := \frac{1}{T}\sum_{i=1}^T\mathcal{M}_s(s\pe{\omega_{\tau}\pe{i}})\,,
    \end{equation}
    where $\omega_{\tau}\pe{1}, \ldots, \omega_\tau\pe{T}$ are all and only the time windows that include the second $\tau$.

    \item\label{item:ensembled_pred} For each second $\tau$ in the time length of the video file, evaluate its estimated highlight possibility rate assigned from the model $\mathcal{M}_e$; i.e., compute the quantity
    \begin{equation}
        \widehat{H}_e(\tau) := \frac{1}{2}\left(\widehat{H}_a(\tau) + \widehat{H}_v(\tau)\right)\,.
    \end{equation}

    \item Classify as a highlight scene with respect to $\mathcal{M}_s$, each second of the video that has value $\widehat{H}_s(\tau)$ greater that a chosen threshold $\epsilon\in (0, 1)$, for each $s=a,v,e$.
    
\end{enumerate}

\subsection{Application Examples}\label{sec:examples_app}

We conclude the section by illustrating two examples, where the models $\mathcal{M}_a,\mathcal{M}_v$, and $\mathcal{M}_e$ (trained with respect to a time window of $k=5$ seconds) are applied to two video clips, with a highlight detection threshold $\epsilon=0.5$. The two video clips have been taken from the Soccernet's public repository (see \cite{Soccernet20218,Soccernet_repo}). In particular, we want to show the efficiency of the ensemble model $\mathcal{M}_e$.

\begin{figure*}[hbt] 
    \centering
    \includegraphics[width=0.75\textwidth]{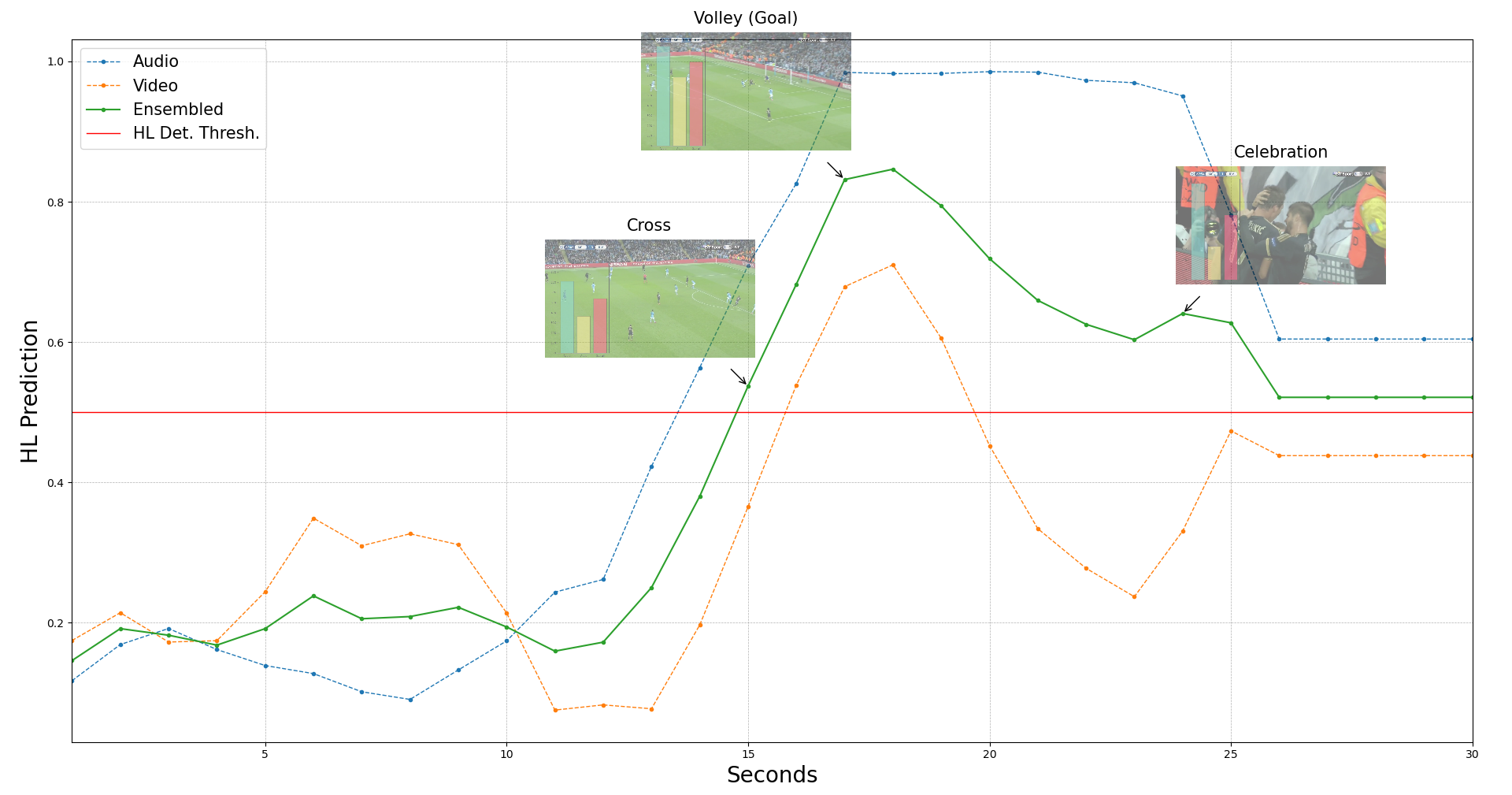} 
    \caption{First clip. Prediction scores of the models; blue $\mathcal{M}_a$ (audio), orange $\mathcal{M}_v$ (video), and green $\mathcal{M}_e$ (ensemble). The red line represents the detection threshold $\epsilon=0.5$ The pictures illustrate the action in the video clip at the seconds pointed out by the arrows.}
    \label{fig:clip2}
\end{figure*}

The first clip is characterized by one goal action. This action is fast ($\sim 3$ seconds), starting with a cross far from the goalpost and ending with a volley. Nonetheless, the ensemble model $\mathcal{M}_e$ successfully identifies a highlight that starts with the cross and lasts until the end of the celebrations. In \Cref{fig:clip2} we illustrate the predictions of all the models, emphasizing the predictions of the ensemble model and adding video frames to explain the analysis better. In particular, looking at \Cref{fig:clip2}, we can see that the audio and video predictions are almost always coherent, except after the goal and before the zoom-in on the celebrating players. In this situation, it is thank to the enthusiasm of the commentator (and the choirs of spectators) that the audio model keeps its predictions high and, therefore, the ensemble model identifies a better highlight (i.e., not only focused on the action, but also on the celebrations).

\begin{figure*}[htb] 
    \centering
    \includegraphics[width=0.85\textwidth]{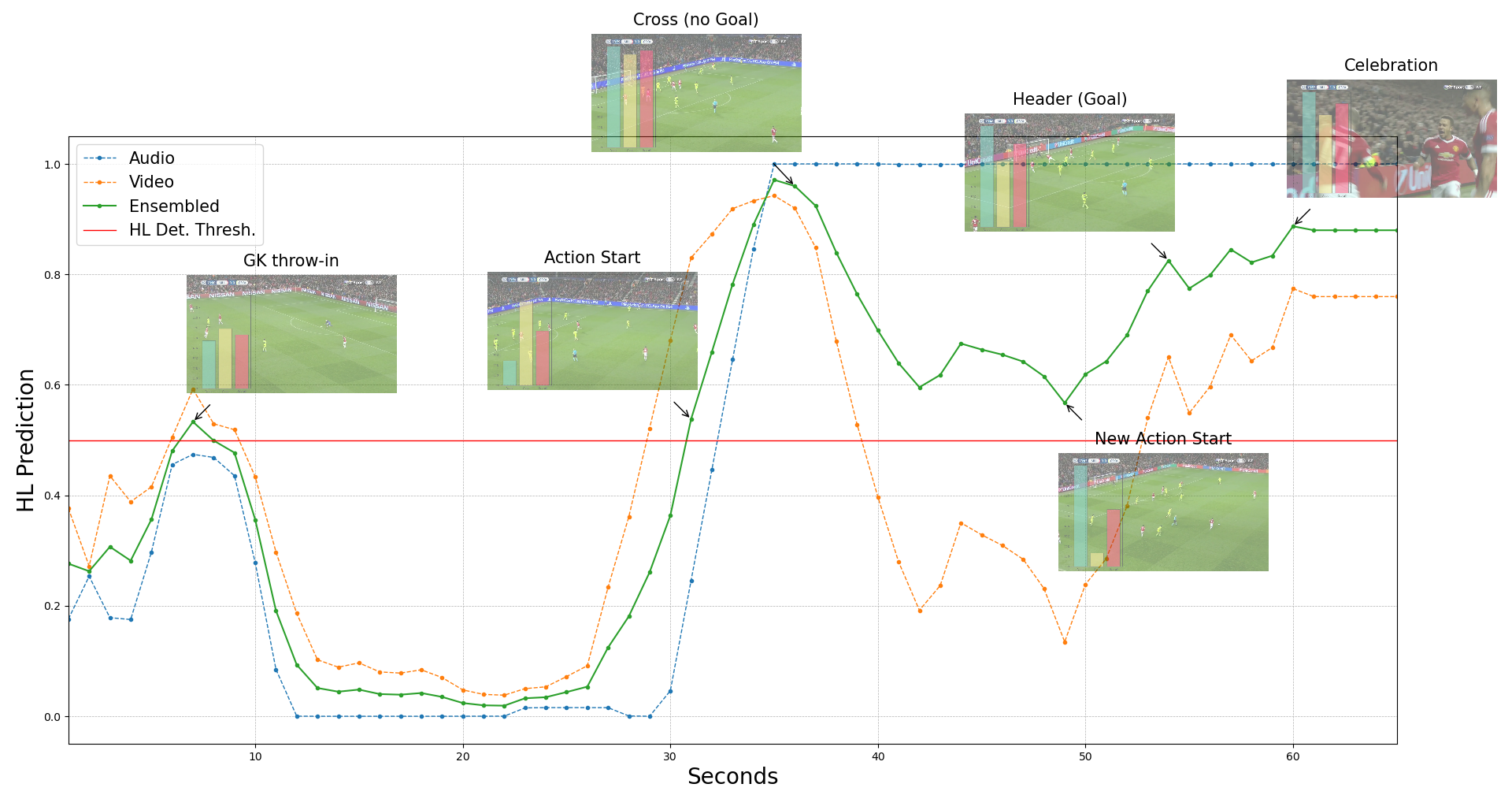} 
    \caption{Second clip. Prediction scores of the models; blue $\mathcal{M}_a$ (audio), orange $\mathcal{M}_v$ (video), and green $\mathcal{M}_e$ (ensemble). The red line represents the detection threshold $\epsilon=0.5$. The pictures illustrate the action in the video clip at the seconds pointed out by the arrows.}
    \label{fig:clip1}
\end{figure*}

The second clip we consider is characterized by two consecutive actions (of the same team): the first action terminates with a failure, but they maintain control of the game and start a second action, scoring a goal. Also in this case, the ensemble model $\mathcal{M}_e$ performs very efficiently, even if a very brief false highlight ($\sim 2$ seconds) is detected during a throw-in of the Goal-Keeper at the beginning of the clip (see ``GK throw-in'' in \Cref{fig:clip1}). This kind of false positive can be easily avoided automatically; e.g., by introducing a criterion of ``minimum length'' for the highlights. Focusing on the two actions contained in this clip, we observe that the ensemble model identifies a unique highlight because there is not a true break between the two. Nonetheless, it is thanks to the ``roar of the crowd'' if the two actions are predicted as one unique highlight; indeed, this sound maintains high the predictions of the audio model, permitting the ensemble model to identify one ``exciting'' highlight (see the rightmost part of the blue line in \Cref{fig:clip1}). On the other hand, the video model, ignoring the sounds, identifies two actions (see the spike of the orange line in \Cref{fig:clip1}); in particular, the video model is responsible for a good choice of the start of the true highlight (see ``Action Start'' in \Cref{fig:clip1}).

\section{Conclusion}\label{sec:conclusion}

In this work, we presented an approach for developing Deep Learning models
able to automatically detect chosen types of scenes in videos. In particular, we focused on the case of highlight detection in recordings of professional football matches. We worked with separated audio and video sources and we formalized the highlight detection problem for both of them. 

We approached the detection from audio sources by developing from scratch a Convolutional NN model that, for each audio chunk of $k$ seconds, predicts if it intersects a highlight time interval or not; specifically, the model receives as inputs the Mel-spectrograms of the audio chunks. 

For detecting highlights from video sources, we proposed to work with greyscale video frames and stack them into tensors, obtaining ``multi-channel'' images that describe video chunks of $k$ second. Then, we developed a Convolutional NN model strongly based on a pre-trained NN model for image classification tasks.

Both the trained models have good performances on the test sets, but they may suffer from false positive/negative predictions, depending on the situation. Therefore, we developed a third ensemble model, with predictions that are the average of the audio model predictions and the video model predictions.

We concluded the work with two experiments. We applied the trained models to two video clips. The results obtained are extremely promising, even if there is still potential for performance improvements.

In conclusion, we presented a methodology for developing efficient and reliable Deep Learning models for fast, cheap, and automated highlight detection in recordings of professional football matches.

Future work may focus on improving the NN architectures and on increasing the amount of training data, for improving the detection performances. Moreover, the development of an automated tool for highlight detection based on these models will be of critical interest; analogously, it would be interesting to extend this methodology to other scene-detection applications.



%




\section*{Acknowledgment}
F.D. acknowledges that this study was carried out within the FAIR-Future Artificial Intelligence Research and received funding from the European Union Next-GenerationEU (PIANO NAZIONALE DI RIPRESA E RESILIENZA (PNRR)–MISSIONE 4 COMPONENTE 2, INVESTIMENTO 1.3---D.D. 1555 11/10/2022, PE00000013). This manuscript reflects only the authors’ views and opinions; neither the European Union nor the European Commission can be considered responsible for them. F.D. acknowledges support from Italian MUR PRIN project 20227K44ME, Full and Reduced order modeling of coupled systems: focus on non-matching methods and automatic learning (FaReX).
M.L. sincerely acknowledges the Deltatre Innovation Lab team for their valuable support and collaboration in shaping the ideas explored in this study. Their encouragement and insights have been instrumental in the development of this work.

\ifCLASSOPTIONcaptionsoff
  \newpage
\fi



\bibliographystyle{IEEEtran}
\bibliography{bibtex/bib/references}
%



%


\begin{IEEEbiographynophoto}{Francesco {Della Santa}}
Research Associate at Politecnico di Torino. He got a Master Degree in Mathematics at University of Florence and a Ph.D. in Pure and Applied Mathematics at Politecnico di Torino. His main scientific interests concern Deep Learning, Surrogate Models, Uncertainty Quantification, and Numerical Optimization.
\end{IEEEbiographynophoto}

\begin{IEEEbiographynophoto}{Morgana Lalli}
She holds a Master's degree in Biomedical Engineering from Politecnico di Torino, along with a second-level Master's in Sports Engineering. Since 2019, she has been part of Deltatre's Innovation Lab, where she contributes to research and development initiatives at the intersection of data science, computer vision, and natural language processing (NLP). Her work focuses on designing data-driven solutions, exploring machine learning methodologies, and investigating the application of large language models (LLMs) and AI agents to enhance the performance and scalability of intelligent systems.
\end{IEEEbiographynophoto}






\end{document}